\documentclass[sigconf]{acmart}

\AtBeginDocument{%
  }

\copyrightyear{2025}
\acmYear{2025}
\setcopyright{acmlicensed}
\acmConference[WWW Companion '25] {Companion Proceedings of the ACM Web Conference 2025}{April 28-May 2, 2025}{Sydney, NSW, Australia.}
\acmBooktitle{Companion Proceedings of the ACM Web Conference 2025 (WWW Companion '25), April 28-May 2, 2025, Sydney, NSW, Australia}
\acmISBN{979-8-4007-1331-6/25/04}
\acmDOI{10.1145/XXXXXX.XXXXXX}

\usepackage{subcaption}
\usepackage{adjustbox}
\usepackage{booktabs}
\usepackage{multirow}
\usepackage{subcaption}
\usepackage{pifont} 
\newcommand{\cmark}{\ding{51}}%
\newcommand{\xmark}{\ding{55}}%
\usepackage{latexsym}
\usepackage[countmax]{subfloat}

\begin{document}

\title{Towards a Robust Framework for Multimodal Hate Detection: \\A Study on Video~\textit{vs.} Image-based Content}

\author{Girish A. Koushik}
\orcid{0000-0002-3875-6503}
\affiliation{%
  \institution{NICE Research,\\ University of Surrey,}
  \city{Guildford}
  \country{United Kingdom}
}
\email{g.koushik@surrey.ac.uk}
\authornote{Corresponding author.}

\author{Diptesh Kanojia}
\orcid{0000-0001-8814-0080}
\affiliation{%
  \institution{NICE Research,\\ University of Surrey,}
  \city{Guildford}
  \country{United Kingdom}}
\email{d.kanojia@surrey.ac.uk}

\author{Helen Treharne}
\orcid{0000-0003-1835-4803}
\affiliation{%
  \institution{Surrey Centre for Cyber Security,\\ University of Surrey,}
  \city{Guildford}
  \country{United Kingdom}}
\email{h.treharne@surrey.ac.uk}

\renewcommand{\shortauthors}{Koushik et al.}

\begin{abstract}
  Social media platforms enable the propagation of hateful content across different modalities such as textual, auditory, and visual, necessitating effective detection methods. While recent approaches have shown promise in handling individual modalities, their effectiveness across different modality combinations remains unexplored. This paper presents a systematic analysis of fusion-based approaches for multimodal hate detection, focusing on their performance across video and image-based content. Our comprehensive evaluation reveals significant modality-specific limitations: while simple embedding fusion achieves state-of-the-art performance on video content (HateMM dataset) with a 9.9\% points F1-score improvement, it struggles with complex image-text relationships in memes (Hateful Memes dataset). Through detailed ablation studies and error analysis, we demonstrate how current fusion approaches fail to capture nuanced cross-modal interactions, particularly in cases involving benign confounders. Our findings provide crucial insights for developing more robust hate detection systems and highlight the need for modality-specific architectural considerations. The code is available at \url{https://github.com/gak97/Video-vs-Meme-Hate}.
  
\textbf{Disclaimer:} This paper discusses publicly available hateful data for academic research only. Examples may contain distasteful content that could be disturbing to readers.
\end{abstract}




\maketitle

\section{Introduction}
The proliferation of hate speech on social media platforms poses a significant social challenge, which requires the development of effective detection and mitigation strategies. While content moderation policies exist, platforms such as BitChute, known for their minimal moderation, often become havens for content that vilifies individuals or communities based on identity, gender, religion, race, nationality, or sexual orientation. Even platforms with stricter moderation policies, like Facebook, X, Instagram, and YouTube, still grapple with the widespread dissemination of hateful content through text, images, and videos. By the time such content gets flagged or removed, it spreads across other platforms too, many of which propagate a high selective bias, leading to incorrect moderation from a particular echo chamber~\cite{cinelli2021echo,gillani2018me}. While polarization and selective bias of social media platforms are not direct contributors to the increase in hateful content, they play a role in propagating such content among groups where it is only likely to be propagated further~\cite{zannettou2018gab}. 

The viral nature of memes, with their ability to convey complex ideas through simple image-text combinations, makes them particularly effective at spreading hateful ideologies across diverse online communities. Similarly, video content, with its engaging audio-visual format, can normalise and amplify hate speech, potentially influencing viewers' attitudes and behaviours. The pervasive nature of such multimodal content underscores the urgent need for robust detection frameworks that can effectively identify and mitigate hate speech across various digital formats.

Existing research has primarily focused on unimodal hate speech detection, with significant progress made in handling individual modalities such as text or image. However, the effectiveness of these approaches across different modality combinations remains poorly understood. This indicates that a holistic framework is needed to address hateful content propagation that encompasses both individual and combined modalities, viz., text, image, audio, or video. Such a framework would also need to be objective while automatically flagging content violating community guidelines or policies.

\begin{figure}[!ht] 
    \centering
    \includegraphics[width=7.5cm]{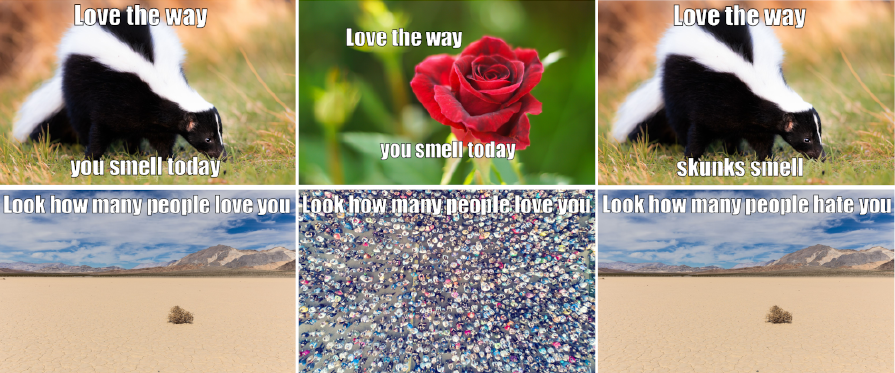}
    \caption{Illustration of benign confounders (not present in the dataset) by~\citet{kiela2020hateful}. (left) meme, (middle) image confounder, (right) text confounder.}
    \label{fig:hatefulmemes_illustration}
\end{figure}

This paper investigates the challenges of multimodal hate speech detection, specifically focusing on the performance discrepancies between video and image-based content. We examine existing resources and different approaches for the task of multimodal hate speech detection, modelling it as a classification problem. Current approaches to multimodal hate speech detection leverage different pre-trained encoders to extract modality-specific unimodal representations or embeddings from a given input, be it an image and a text segment, or video, speech segment, and text.  We conduct a systematic analysis of fusion-based approaches, evaluating their effectiveness on two distinct datasets: HateMM~\cite{das2023hatemm}, representing video content, and the Hateful Memes Challenge dataset (HMC)~\cite{kiela2020hateful}, representing image-text combinations. Our comprehensive evaluation reveals significant modality-specific limitations in current methods. Notably, while simple embedding fusion achieves state-of-the-art performance on the video-based HateMM dataset, it struggles to capture the complex semantic relationships present in the image-text-based Hateful Memes dataset due to the necessity of a combined understanding of image and text.~\citet{kiela2020hateful} indicate that a crucial characteristic of the challenge with image-based datasets is the inclusion of ``benign confounders'', illustrated in Figure~\ref{fig:hatefulmemes_illustration}, which counter the possibility of models exploiting unimodal priors.

We analyse the performance of various approaches and present a comprehensive evaluation based on two existing datasets differing over available modalities: image \& text; visual, speech \& text. Existing research on multimodal hate speech detection leverages datasets with varying modalities~\cite{chhabra2023multimodal, arya2024multimodal, aziz2023csecu}. However, to the best of our knowledge, no existing work evaluates the performance of their approach on differing modalities despite commonalities in methods applied to the task, which is essential for building a robust framework. This comparison enables researchers to determine whether existing approaches to multimodal fusion serve as an appropriate foundation for a holistic framework for mitigating hate speech. 
The contributions of our work can be summarised as follows:
\begin{itemize}
    \item Comprehensive cross-modal analysis that demonstrates how fusion approaches perform differently across video and image-based hate content, revealing modality-specific limitations in current methods.
    \item Systematic evaluation showing a 9.9\% points F1-score improvement on video content (HateMM dataset) using simple embedding fusion, while demonstrating its limitations on complex image-text relationships in the HMC dataset.
    \item Detailed ablation studies and error analysis providing concrete insights into where and why current fusion approaches succeed or fail across different modality combinations.
\end{itemize}

\section{Related Work}
Detection of hate speech has evolved from text-focused to multimodal methods, reflecting the changing nature of online hate speech that includes visuals. This review highlights the limitations of current methods and the need for robust approaches across modalities. Additionally, we categorise existing datasets by available modalities.


\subsection{Text-based Hate Speech Detection}
Research on hate speech detection initially centred on textual content. Various data sets such as Hate Speech~\citep{de2018hate}, ETHOS~\citep{mollas2020ethos}, Twitter Hate Speech~\citep{waseem2016hateful}, and HateXplain~\citep{mathew2021hatexplain, chhabra2023literature} support this work. Often sourced from Twitter, YouTube, and Reddit, these datasets are instrumental in training and evaluating text-based models. ETHOS\footnote{\url{https://hatespeechdata.com}} provides leaderboards for binary and multi-label classification, with BiLSTM and BERT excelling in binary tasks, and BiLSTM with attention in multi-label tasks~\citep{mollas2020ethos}. HateXplain, derived from Twitter and Gab, adds value by including rationale labels explaining hate speech. A BERT model fine-tuned for these labels achieves top performance with AUROC scores~\citep{mathew2021hatexplain}. However, text-only methods overlook the multimodal nature of online hate speech, where meaning arises from textual and visual elements, leading to potential detection inaccuracies.


\subsection{Multimodal Hate Speech Detection}
Recognising the limitations of text-based approaches, recent research has shifted towards multimodal hate speech detection, aiming to leverage information from both text and other modalities like images, audio and video. This shift has been accompanied by the development of new datasets that include multiple modalities.

\subsubsection{Multimodal Datasets}
Existing datasets differ in terms of modality-specific data and content sources. For instance,~\citet{gomez2020exploring} released the MMHS150K dataset encouraging early research on multimodal hate speech detection, closely followed by the introduction of the HMC dataset~\citep{kiela2020hateful}, specifically designed to challenge models with instances where hate speech is conveyed through the combination of image and text, even if neither modality is hateful in isolation. The inclusion of "benign confounders" in the Hateful Memes dataset, where either the image or text is altered to create a hateful/non-hateful meme, further challenges models to understand the nuanced interplay between modalities. Facebook improved upon HMC by providing fine-grained labels on the \textit{type of attack} and \textit{target of hate} labels~\citep{mathias2021findings}. These fine-grained labels enabled~\citet{hee2023decoding} to extend the dataset's annotations, adding the underlying reasoning behind the assigned hate labels, thus providing a deeper understanding of the hateful content.~\citet{das2023hatemm} introduced the first publicly available video-based hate speech dataset referred to as HateMM, which includes visuals and audio features. Beyond these, several other datasets have emerged, focusing on specific topics or events, including MultiOFF~\citep{suryawanshi2020multimodal} related to the $2016$ US Presidential Elections, the HarMeme dataset~\citep{pramanick2021detecting} related to COVID-19, and CrisisHateMM dataset~\citep{bhandari2023crisishatemm} on Russia-Ukraine conflict. 

\subsubsection{Challenges in Multimodal Hate Speech Datasets}
While most of the above data instances can make underlying hate apparent through an individual modality (either text or image), HMC is an exception as it requires foundation models to jointly understand both text and image to a greater extent due to benign confounders. On the other hand, HateMM contains more information as a continuous stream, adding necessary context for input. HateMM data curation relies on automatic speech recognition (ASR) to obtain the transcripts for its videos. The absence of gold-standard transcripts is an underlying challenge for such datasets. This challenge is further elevated by the content source of data instances. Our analysis indicates that many videos are present in the form of songs containing either music or visual-only instances, adding more complexity to the task. Further, only HMC has existing benchmark approaches\footnote{\url{https://paperswithcode.com/sota/meme-classification-on-hateful-memes}}, some of which we discuss below.

\subsection{Multimodal Fusion Approaches}
Early multimodal work such as MMHS150K~\citep{gomez2020exploring} used simple concatenation of image and text features. The introduction of the HMC dataset~\citep{kiela2020hateful} revealed the limitations of such approaches when dealing with benign confounders where neither the text nor the image alone is hateful, but the combination is. Several subsequent approaches have attempted to address this challenge through more sophisticated architectures and fusion mechanisms.

\subsubsection{Visio-Textual Fusion}
Recent approaches have attempted to address this through sophisticated architectures: HateCLIPper~\citep{kumar2022hate} fine-tunes Contrastive Language Image Pre-training (CLIP)~\citep{radford2021learning} projections, while ~\citet{mei2023improving} improves the HateCLIPper approach by aligning embeddings from the same class, dynamically retrieving them and training them with a contrastive loss along with cross-entropy loss.~\citet{burbi2023mapping} also utilise the CLIP model but disentangle the representations before carrying out textual inversion, \textit{i.e.,} image and text features are fused using a combiner module and passed to an MLP for binary classification. Similarly, a fine-tuned Flamingo~\citep{alayrac2022flamingo} also performs well on the HMC dataset. Modularised networks have been employed for few-shot learning, improving detection performance on smaller datasets~\citep{cao2024modularized}. Multi-scale visual kernels combined with knowledge distillation architecture have also been utilised to enhance robustness and accuracy~\citep{chhabra2023multimodal}. PALI-X-VPD~\citep{hu2023visual} employs a $55B$ parameter language model with access to code generation and external tools such as object detection, visual question answering, optical character recognition, etc, eventually aggregating this information with the help of chain-of-thought. While these approaches achieve strong performance ($AUROC > 0.85$), they require significant computational resources and may not generalise to other modality combinations. More recently, zero-shot evaluation of large multimodal models (LMMs) such as LLaVA-1.5, BLIP-2, Evolver, etc. has been performed on this dataset however they still fall short of the SoTA approaches~\citep{huang2024evolver}.

\subsubsection{Video-based Fusion}
Video hate detection presents additional challenges due to temporal dependencies and the need to integrate audio features. The HateMM dataset~\citep{das2023hatemm} pioneered work in this direction, employing pre-trained encoders for text, image and audio and obtaining sequential (temporal) information with LSTM~\citep{hochreiter1997long} and aggregating them with simple embedding fusion. MultiHateClip~\citep{wang2024multihateclip} extended this to multilingual content but maintained similar fusion strategies. Notably, approaches that succeed on video content often fail on static image-text pairs, suggesting fundamental limitations in current fusion methods. While most of the work discussed above focuses primarily on performance improvement, none discuss a common framework to combat modality-agnostic hate content. Our work aims to bridge this gap by conducting a comparative analysis of fusion approaches across video and image-based content, shedding light on their strengths and weaknesses in different multimodal contexts.


\section{Methodology}
Our analysis focuses on two fundamental questions: (1) how do fusion approaches perform across different modality combinations, specifically video-based content and meme-based content, and (2) what factors influence the success or failure of these approaches in detecting hate speech? To answer these questions, we conduct a systematic evaluation using two representative datasets that exemplify different multimodal challenges: HateMM for video content and HMC for image-text combinations. 

\subsection{Datasets} 
We selected two datasets that pose distinct challenges for multimodal hate speech detection:

\paragraph{HateMM (Video-based)}: This dataset consists of $1083$ labelled videos extracted from the BitChute platform using hate-related lexicons, with $431$ categorised as hate and $652$ as non-hate. The dataset presents challenges related to temporal dependencies, variable-length content ($1$ second to $1.5$ hours), and missing modalities ($48$ videos lack audio). Moreover, the ASR-generated captions from the audio contain inaccuracies due to noisy audio or videos containing music. While some videos lack audio, their visuals still convey hate, emphasizing the importance of utilising these data points\footnote{In cases of data samples missing any modality, we pass zero tensors for those modalities and proceed to fusion}. Given the dataset's modest size of approximately 1000 samples, removing videos could compromise the analysis. We employ a split of $779$ videos for training, $87$ for validation, and $217$ for testing.

\paragraph{Hateful Memes Challenge (HMC) (Image-Text-based)}: This dataset contains $8.5k$ train, $500$ validation and $1k$ test images released as part of the Hateful Memes Challenge~\citep{kiela2020hateful}. It includes instances with benign confounders to challenge unimodal priors in approaches depending heavily on either text or image embeddings. This dataset tests a fusion method's ability to capture complex cross-modal interactions. Akin to offensive language identification, the detection of hateful content is also subjective. Many instances of hate content discriminate against a certain gender, race, religion, or nationality, whereas others may not find it hateful~\citep{weerasooriya2023vicarious}. Therefore, data annotation should have a consensus in the form of a strong inter-annotator agreement. Cohen's Kappa for this agreement is $62.5$ for the HateMM dataset~\citep{das2023hatemm} and $68.4$ for the HMC dataset~\citep{kiela2020hateful} which \textit{only shows moderate agreement among annotators} for both datasets further underlining a challenge in objectivity, for a unified multimodal hate detection framework. 

\subsection{Preprocessing and Feature Extraction} \label{subsec:preprocess}

\subsubsection{HateMM Preprocessing}
We improved upon the original HateMM data preprocessing pipeline by replacing the Vosk\footnote{\url{https://alphacephei.com/vosk/models}} ASR tool with the Whisper tiny model for generating video transcripts. This change resulted in significantly improved transcription quality, as demonstrated in Appendix A, due to Whisper's superior ability to handle noisy audio and variations in speech. Due to the absence of any reference text, we are only able to analyse and report qualitative improvements as shown in Appendix~\ref{sec:appendix_tq}.

\subsubsection{Feature Extraction}

\paragraph{Text Models}  
We used the widely adopted BERT~\citep{devlin2018bert} (bert-base-uncased) model to generate contextualized text embeddings. BERT's pre-training on masked language modelling and next-sentence prediction allows it to capture nuanced semantic relationships within the text. We extract the [CLS] embedding from the final layer as a $768-$ dimensional representation of the text. HateXplain~\citep{mathew2021hatexplain} (HXP), on the other hand, is a BERT variant specifically fine-tuned on the HateXplain dataset, which provides word and phrase-level annotations of hate speech targets and rationales. This fine-tuning makes HateXplain particularly well-suited for extracting text embeddings relevant to hate speech detection. HateXplain also generates $768-$ dimensional text embeddings.
For this multimodal investigation, we additionally leverage text embeddings from various multimodal models such as CLIP~\citep{radford2021learning}, Contrastive Language-Audio Pre-training (CLAP)~\citep{elizalde2023clap}, which are image-text and audio-text models, respectively, pre-trained with a contrastive learning objective. We extract text embeddings using these two models and use them for fusion. While CLIP has a dimension of $512$ for its text encoder, CLAP generates embeddings with a dimension of $768$. We use these encoder models to extract the embeddings for the HateMM and HMC datasets. The text encoders used for our experiments range from $63M$ to $140M$ parameters.

\paragraph{Audio Models} 
We comprehensively examine the effect of different audio feature encoding methods. To reproduce the existing work~\citet{das2023hatemm}, we use MFCC and AudioVGG19 models. In addition to this, we propose using Wav2Vec 2.0~\citep{baevski2020wav2vec} for its self-supervised speech representations and CLAP~\citep{elizalde2023clap} for its ability to jointly represent audio and text in the embedding space. MFCC provides a power spectrum representation of the audio based on the linear cosine transform on a non-linear mel-frequency scale~\citep{davis1980comparison}. MFCC representations are extracted using Librosa\footnote{\url{https://librosa.org/doc/latest/index.html}} to obtain a 40-dimensional vector. AudioVGG, the VGG19 model fine-tuned on audio style transfer, takes waveforms of audio files as input and produces a 1000-dimensional vector representation \citep{grinstein2018audio}~\citep{simonyan2014very}. Also, CLAP allows the processing of audio files greater than $10$ seconds without chunking. However, the audio files had to be chunked into $30$-second segments for Wav2vec $2.0$ audio embeddings. The audio encoders used for our experiments range from $80M$ to $94M$ parameters except for MFCC, which uses Fourier and cosine transforms for feature extraction.

\paragraph{Vision Models}  
Originally,~\citet{das2023hatemm} used Vision Transformer (ViT)~\citep{dosovitskiy2020image} as image encoder and extract $100$ frames ($1$ frame per second) from each video to obtain image embeddings.
In addition to ViT, we leverage CLIP~\citep{radford2021learning} for its ability to jointly represent image and text in the embedding space and DINOv2~\citep{oquab2023dinov2} for its powerful self-supervised image understanding. DINOv2 improves upon ViT by knowledge distillation, generating smaller models, and utilising local and global views of the image patches to improve image understanding. The embeddings of these models are also passed through the LSTM for the video dataset for sequential temporal information. The models yield a $768$ (for ViT \& CLIP) and $384$ (for DINOv2) dimension embedding for each image. All the above vision models are used to extract the image embeddings on the HMC dataset as well. The image encoders used for our experiments range from $22M$ to $86M$ parameters.

\begin{table}[!ht]
    \centering
    \renewcommand{\arraystretch}{1.5} 
    \resizebox{\columnwidth}{!}{%
    \begin{tabular}{c c c}
    \toprule
     \textbf{Model} & \textbf{Embeddings} & \textbf{Dataset} \\ 
    \midrule
     BERT (bert-base-uncased) & Text & HatefulMemes, HateMM \\ 
     HateXplain & Text & HatefulMemes, HateMM \\ 
     CLIP (clip-vit-base-patch32) & Image, Text & HatefulMemes, HateMM \\ 
     ViT (vit-base-patch16-224-in21k) & Image & HatefulMemes, HateMM \\ 
     DINOv2 (dinov2-small) & Image & HatefulMemes, HateMM \\ 
     CLAP (clap-htsat-unfused) & Audio, Text & HateMM \\ 
     MFCC & Audio & HateMM \\ 
     AudioVGG19 & Audio & HateMM \\ 
     Wav2Vec2 (wav2vec2-base-960h) & Audio & HateMM \\ 
    \bottomrule
    \end{tabular}%
    }
    \caption{Encoder models for different modalities}
    \label{embeddings_table}
\end{table}

\subsection{Fusion Approaches}

We evaluate two distinct fusion paradigms simple embedding fusion (hereon denoted as Simple Fusion) and modality order-aware fusion (hereon denoted as MO-Hate):

\subsubsection{Simple Embedding Fusion}
This approach combines modality-specific embeddings through concatenation or element-wise operations. For a given input with $m$ modalities, the fused representation $F$ is computed as:
For concatenation: \(F = [E_1; E_2; ...; E_m]\)
For element-wise product: \(F = E_1 \odot E_2 \odot ... \odot E_m\)
where $E_i$ represents embeddings from modality $i$. This approach assumes modalities contribute independently to the final representation.
~\citet{sai2022explorative} describes \textit{concatenation} and \textit{product rule} (or element-wise product) as early fusion methods whereas, \textit{weighting techniques} and \textit{rules learned from training on probabilities} as late fusion methods. \textit{Concatenation} involves concatenating text, audio, and image embeddings for the video dataset (as shown in Appendix~\ref{sec:appendix_arch} Figure~\ref{fig:baseline}) and image + text for the HMC dataset. However, for \textit{product rule}, an element-wise product is computed amongst the embeddings and then passed to a classifier for predictions. Late fusion techniques such as \textit{weighting techniques} and \textit{rules learned from training on probabilities} entail passing the individual embeddings through the classifier, dynamically assigning weights to the modality and then sending them through another classifier, which learns from the probabilities of the first classifier~\citep{sai2022explorative}.

\subsubsection{Modality Order-Aware Fusion (MO-Hate)}

This approach, inspired by~\citet{tomar2023your} work on multimodal sarcasm detection, incorporates sequential dependencies between modalities using a modified BART~\citep{lewis2019bart} architecture. The MO-Hate architecture addresses three key dimensions of multimodal fusion: semantic information within modalities, contextual relationships across modalities, and temporal dependencies. Mathematically, this can be expressed as:
\[F_{total} = \sum_{j \in \{text, image, audio\}} [ \alpha (w_j F_{semantic}) + \beta (w_j F_{contextual})\] 
\[+ \gamma (F_{temporal})]\]
where $\alpha, \beta, \gamma$ are coefficients, $w_j$ represents the modality inputs. $F_{semantic}$ captures content meaning across each modality, $F_{contextual}$ addresses context across modalities, and $F_{temporal}$ accounts for evolution over time. This is for a dataset containing temporal information in the form of audio or video. In the case of visio-textual datasets, the $F_{temporal}$ factor would be absent.

The MO-Hate model first encodes the text input using the BART encoder, generating a $768$-dimensional embedding. It then uses a modality fusion network to sequentially process audio and visual inputs, generating context-infused text embeddings at each step. A dense layer transforms the modality sequence length to match the text sequence length. The audio query vector (context), text key, and value vectors are computed, undergo element-wise multiplication, and the product is used as the contextualised input for the next layer with visual input fusion (as shown in Appendix~\ref{sec:appendix_arch} Figure~\ref{fig:mo-hate}). This process allows each modality to influence the representation of subsequent modalities through an attention mechanism. For modalities i and j, the fusion process computes: \(Q = W_iE_i, K = W_jE_j, V = W_vE_j\)
\[\text{Attention}(Q, K, V) = \text{softmax}\left(\frac{QK^T}{\sqrt{d_k}}\right)V\]
where $W_i$, $W_j$, and $W_v$ are learned parameters, $E_i$ and $E_j$ are input embeddings and \(d_k\) is the dimension of the key vector.

This formulation guides our implementation where we use attention mechanisms to learn these relationships. The semantic component is handled through modality-specific encoders, contextual relationships through cross-modal attention, and temporal aspects through sequential processing of modalities.

\begin{table*}[!htb]
    \begin{subtable}{0.58\textwidth}
      \centering
        \renewcommand{\arraystretch}{1.2}
        \resizebox{\textwidth}{!}{
        \begin{tabular}{c|c c c c c}
        \toprule
        & \textbf{Models} & \textbf{F1 (M)} & \textbf{P (M)} & \textbf{R (M)} & \textbf{Acc} \\
        \midrule
        \multirow{4}{*}{\rotatebox{90}{Existing}}
        & BERT + MFCC + ViT~\citep{das2023hatemm} & {0.749} & 0.742 & 0.758 & 0.798 \\
        & HXP + MFCC + ViT~\citep{das2023hatemm} & 0.720 & 0.718 & 0.726 & 0.777 \\
        & BERT + AVGG19 + ViT~\citep{das2023hatemm} & 0.718 & 0.723 & 0.719 & 0.755 \\
        & HXP + AVGG19 + ViT~\citep{das2023hatemm} & 0.707 & 0.714 & 0.712 & 0.767 \\
        \midrule
        \multirow{3}{*}{\rotatebox{90}{Sim Fusion}} 
        & HXP + CLAP + ViT (Concat) & 0.823 & 0.803 & 0.765 & 0.832 \\
        & CLAP Text + CLAP Audio + CLIP (Concat) & 0.802 & 0.788 & 0.741 & 0.811 \\
        & HXP + CLAP + CLIP (Concat) \textbf{(HCC1)} & \textbf{0.848} & \textbf{0.840} & 0.800 & \textbf{0.854} \\
        \midrule
        \multirow{3}{*}{\rotatebox{90}{MO-Hate}} 
        & BART $\rightarrow$ Wav2Vec2 $\rightarrow$ DINOv2 & {0.821} & 0.822 & \textbf{0.820} & 0.820 \\
        & Wav2Vec2 $\rightarrow$ ViT $\rightarrow$ Text (BART) & 0.794 & 0.794 & 0.794 & 0.794 \\
        & BART $\rightarrow$ CLAP $\rightarrow$ DINOv2 \textbf{(BCD1)} & \underline{0.821} & 0.821 & 0.820 & 0.820 \\
        \bottomrule
    \end{tabular}
    }
    \caption{Comparison with existing methods. For MO-Hate results, modality order is indicated using arrows. Best scores are in boldface, and the next best are underlined. HCC1 is best performing approach for baseline, while BCD1 is best from MO-Hate.}  
    \label{hatemm_table}
    \end{subtable}%
    \quad
    \begin{subtable}{.4\textwidth}
      \centering
      \resizebox{\textwidth}{!}{
        \begin{tabular}{c|c c c c c c c}
        \toprule
        & \textbf{T} & \textbf{A} & \textbf{V} & \textbf{F1 (M)} & \textbf{P (M)} & \textbf{R (M)} & \textbf{Acc} \\
        \midrule
        \multirow{6}{*}{\rotatebox{90}{Simple Fusion}} 
        & \cmark & \xmark & \xmark & 0.786 & 0.733 & 0.776 & 0.791 \\
        & \xmark & \cmark & \xmark & 0.730 & 0.739 & 0.600 & 0.747 \\
        & \xmark & \xmark & \cmark & 0.738 & 0.698 & 0.682 & 0.747 \\
        & \cmark & \cmark & \xmark & 0.778 & 0.791 & 0.670 & 0.791 \\
        & \cmark & \xmark & \cmark & \textbf{0.819} & 0.795 & 0.776 & 0.825 \\
        & \xmark & \cmark & \cmark & 0.803 & 0.780 & 0.753 & 0.810 \\
        \midrule
        \multirow{6}{*}{\rotatebox{90}{MO-Hate}} 
        & \cmark & \xmark & \xmark & 0.785 & 0.784 & 0.785 & 0.791 \\
        & \xmark & \cmark & \xmark & 0.614 & 0.631 & 0.614 & 0.645 \\
        & \xmark & \xmark & \cmark & 0.717 & 0.719 & 0.716 & 0.728 \\
        & \cmark & \cmark & \xmark & \textbf{0.831} & 0.829 & 0.833 & 0.835 \\
        & \cmark & \xmark & \cmark & 0.806 & 0.832 & 0.798 & 0.820 \\
        & \xmark & \cmark & \cmark & 0.802 & 0.807 & 0.798 & 0.810 \\
        \bottomrule
    \end{tabular}
    }
    \caption{Ablation highlighting pair-wise modality contributions.}
    \label{modality_table}       
    \end{subtable} 
    \caption{(a) Performance on the HateMM Dataset (b) Ablation study results. M: macro, P: precision, R: recall, Acc: accuracy.}
\end{table*}

\subsection{Experimental Setup}
All models are obtained from HuggingFace and used to extract the embeddings as described above and summarised in Table~\ref{embeddings_table}. For HMC, we modify MO-Hate, to eliminate the audio encoding by passing zero tensors. For simple fusion code, the available embeddings were either concatenated or element-wise product was computed. The experiments are conducted using two $24GB$ $A5000$ GPUs. All experiments on the HateMM and HMC datasets are executed with a batch size of $32$, a learning rate of $1e-4$, binary cross-entropy loss and for $20$ epochs. The runtime for all the HateMM experiments was nearly $30$ minutes per run, totalling $10$ GPU hours whereas for HMC the average runtime was $90$ minutes, totalling $30$ GPU hours. To avoid overfitting, a $20$\% dropout and early stopping are used for experiments with both baseline and MO-Hate architectures on HMC.
\paragraph{Evaluation Metrics} 
The reported metrics for the HateMM dataset are macro-F1 score along with precision, recall and accuracy. On the HMC dataset, even though the widely reported metric is the Area Under the Receiver Operating Characteristic Curve (AUROC), we also report accuracy, precision, recall and F1. We treat macro-F1 as the primary metric for HateMM, and AUROC for HMC datasets. 

\section{Experimental Results}
We present our analysis in three parts: (1) comparative performance of fusion approaches on the HateMM (video) and HMC (image-text) datasets, (2) detailed ablation studies showing the contribution of individual modalities, and (3) qualitative analysis of success and failure cases to provide insights into model behaviour. This structured analysis reveals important insights about the effectiveness of different fusion approaches across modality combinations.

\subsection{Performance on Video Content}
Table~\ref{hatemm_table} presents the results of our experiments on the HateMM dataset. Our best model \textbf{HCC1} achieves SoTA F1 of $0.848$, a significant improvement of $9.9$\% points over the best previously reported results~\citep{das2023hatemm}. This improvement can be attributed to three key factors:
\begin{itemize}
    \item \textbf{Enhanced Transcription Quality}: Replacing the original Vosk ASR tool with the Whisper tiny model led to substantial improvements in the quality of the video transcripts, as illustrated in Appendix~\ref{sec:appendix_tq}.
    \item \textbf{Effective Embedding Fusion}: Simple concatenation of embeddings proved highly effective for this dataset.
    \item \textbf{Strong Pre-trained Encoders}: The use of powerful pre-trained encoders for each modality contributed to the overall performance.
\end{itemize}

These results demonstrate that for video content, where modalities are naturally synchronised, simple fusion strategies can effectively capture cross-modal relationships and achieve strong performance. As shown in Table~\ref{hatemm_table}, the existing architectures evaluated by~\citet{das2023hatemm} achieved a maximum F1 score of $0.749$. Our model (HCC1) significantly outperforms this baseline, demonstrating the effectiveness of our chosen encoders and fusion approach.

Within the MO-Hate architecture, HateXplain and BART models were used to obtain text embeddings on this transcript. In addition to the existing encoders, we investigated the use of Wav2Vec2 and CLAP as audio encoders and CLIP and DINOv2 as image encoders. Compared to the text-first model (\textbf{BCD1}), the audio-first model (second MO-Hate model) does not perform as well. While CLAP performs well on both architectures, it is slightly worse when it comes to lengthy videos as CLAP is originally trained on snippets of audio with text captions. However, Wav2Vec2 performs better with long-form videos owing to its powerful audio representations for length audios. Among the image encoders CLIP and DINOv2, CLIP performs worse when the visual is misleading the audio and thus text or vice versa. The performance between BART and HXP text encoders are quite similar with an edge for BART owing to the better fusion process in MO-Hate. We conducted exhaustive experimentation with different encoder and modality combinations. Additional experimental results have been discussed in Appendix~\ref{sec:appendix_results}, Table~\ref{hatemm_table_appendix}.

\subsection{Performance on Image-Text Content}
Table~\ref{hateful_memes_table} presents the results of our experiments on the HMC dataset. In contrast to the video results, both simple fusion and MO-Hate approaches struggle to achieve high performance on this dataset. Our best-performing model (\textbf{MBD1}) achieves an AUROC of only $0.628$, significantly below state-of-the-art approaches like PALI-X-VPD~\citep{hu2023visual}, which reports an AUROC of $0.892$. This performance gap highlights the fundamental limitations of current fusion approaches when dealing with complex image-text relationships, particularly in the presence of benign confounders.
As indicated in Table~\ref{hateful_memes_table}, existing methods such as RGCL-HateCLIPper~\citep{mei2023improving}, fine-tuned Flamingo~\citep{alayrac2022flamingo}, and Hate-CLIPper-Align~\citep{kumar2022hate} achieve AUROCs ranging from $0.858$ to $0.867$. Our model's significantly lower performance underscores the challenges posed by this dataset. The MO-Hate model (\textbf{MBD1}) utilising BART and DINOv2 achieved an AUROC of $0.628$. While this represents a slight improvement over simple fusion approaches and the zero-shot performance of VLMs, it still falls far short of the state-of-the-art. This suggests that the sequential processing and attention mechanisms in MO-Hate are not sufficient to fully capture the complex semantic relationships in memes. We conducted extensive experimentation with different encoder and fusion combinations, as detailed in Appendix~\ref{sec:appendix_results}, Table~\ref{hatememes_table_appendix}.

\begin{table*}[!ht]
    \centering
    \renewcommand{\arraystretch}{1.15}
    \begin{tabular}{c|c c c c c c}
        \toprule
        & \textbf{Models} & \textbf{AUROC} & \textbf{F1 (M)} & \textbf{P (M)} & \textbf{R (M)} & \textbf{Acc} \\
        \midrule
        \multirow{4}{*}{\rotatebox{90}{0-shot VLMs}}
        & OpenFlamingo (7B) \citep{awadalla2023openflamingo} & 0.570 & - & - & - & 0.564 \\
        & LLaVA-1.5 (13B) \citep{liu2024improved} & 0.618 & - & - & - & 0.614 \\
        & InstructBLIP (13B) \citep{liu2024visual} & 0.596 & - & - & - & 0.601 \\
        & Evolver (13B) \citep{huang2024evolver} & 0.603 & - & - & - & 0.604 \\
        \midrule
        \multirow{4}{*}{\rotatebox{90}{Existing}}
        & \textbf{PALI-X-VPD}~\citep{hu2023visual} & \textbf{0.892} & - & - & - & - \\
        & RGCL-HateCLIPper~\citep{mei2023improving} & 0.867 & - & - & - & 0.788 \\
        & Flamingo - fine-tuned~\citep{alayrac2022flamingo} & 0.866 & - & - & - & - \\
        & Hate-CLIPper - Align~\cite{kumar2022hate} & 0.858 & - & - & - & - \\
        \midrule
        \multirow{3}{*}{\rotatebox{90}{Sim Fusion}} 
        & HXP + CLIP (Concat) & 0.615 & 0.557 & 0.643 & 0.492 & 0.617 \\
        & CLIP Text + CLIP Image (Concat) & 0.606 & 0.531 & 0.644 & 0.451 & 0.609 \\
        & CLIP Text + CLIP Image (EW Product) & 0.591 & 0.467 & 0.660 & 0.361 & 0.596 \\
        \midrule
        \multirow{2}{*}{\rotatebox{90}{\small MO-H}} 
        & BART $\rightarrow$ CLIP & 0.618 & 0.608 & 0.637 & 0.622 & 0.622 \\
        & BART $\rightarrow$ DINOv2 \textbf{(MBD1)} & \underline{0.628} & 0.619 & 0.645 & 0.631 & 0.631 \\
        \bottomrule
    \end{tabular}
    \caption{Performance obtained on HMC dataset. Modality-order for MO-Hate (MO-H) is indicated using arrow. MO-H: MO-Hate, EW: element-wise. The best score from our experiments is underlined. \textbf{MBD1} is the best-performing model among our experiments. M: macro, P: precision, R: recall, Acc: accuracy.}
    \label{hateful_memes_table}
\end{table*}

\subsection{Ablation Study}
To understand the contribution of individual modalities, we conducted ablation studies on both datasets. Table~\ref{modality_table} presents the results of the ablation study on the HateMM dataset. Using only text embeddings achieved a strong F1 score of $0.786$, suggesting that the text modality carries significant information for hate speech detection in this dataset. Combining audio and visual modalities achieved an F1 score of $0.803$, indicating complementary information between these modalities. The best performance ($F1: 0.848$) was achieved when all three modalities were used, highlighting the importance of multimodal fusion. Note that with the absence of visual modality, MO-Hate performs better than \textbf{BCD1} because the model has ignored many instances of benign visuals and focussed only on the hateful audio.

Since memes require a combined understanding of both text and image modalities, conducting a traditional ablation study by removing modalities was not feasible. However, the results in Table~\ref{hateful_memes_table}, using HXP text and image embeddings, provide some insights. The low performance ($AUROC < 0.62$) even with these powerful unimodal embeddings suggests that capturing cross-modal relationships is crucial for this dataset. These patterns suggest that current fusion approaches effectively combine complementary information in synchronised modalities (video) but struggle with complex semantic relationships (memes).

\subsection{Qualitative Analysis}
To gain further insights into model behaviour, we conducted a qualitative analysis of success and failure cases for some instances manually picked from the datasets. 
\textbf{HateMM Success Cases}: The models generally performed well when the hate speech was explicitly expressed in at least one modality. For instance, in videos with hateful audio content, the models successfully detected hate speech even if the visual content was benign.
\textbf{HateMM Failure Cases}: The models struggled with cases where the hate speech was conveyed through subtle cues or required an understanding of the broader context. For example, in videos with sarcastic or implicit hate speech, the models often fail to detect the underlying hateful intent. Moreover, as seen in Table~\ref{hatemm_table}, both \textbf{BCD1} and \textbf{HCC1} seem to be struggling with videos that have misleading (benign) visuals (e.g., a cartoon of trains going around) while the hate-infused audio and text are incomplete or contain noise. Moreover, the MO-Hate model performs better than the Baseline model when only visual data is available. In summary, the models perform best when all three modalities are present or when noise-free audio is available with a good-quality transcript.

\begin{figure}
    \centering
    \subfloat[\centering True Label: \underline{Not Hateful}, Prediction: Not Hateful]{{\includegraphics[width=0.49\columnwidth]{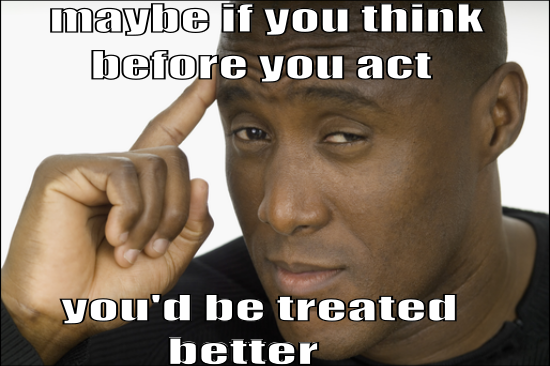} }}%
    \subfloat[\centering True Label: \underline{Hateful}, Prediction: Not Hateful]{{\includegraphics[width=0.49\columnwidth]{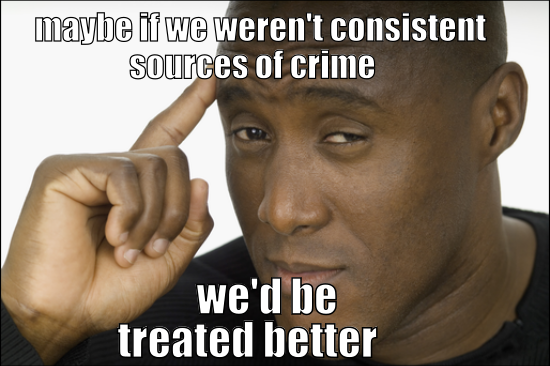} }}%
    \caption{Sample predictions from MBD1.}%
    \label{fig:hatefulmemes_error_analysis}%
\end{figure}

\textbf{Hateful Memes Success Cases}: The models successfully detected hate speech in memes where the hateful content was relatively straightforward, such as those with explicit slurs or derogatory imagery.
\textbf{Hateful Memes Failure Cases}: The models struggled with memes containing benign confounders, where a change in text or image alters the meaning of the meme. For example, as shown in Figure~\ref{fig:hatefulmemes_error_analysis}, a meme with a seemingly innocuous image paired with a hateful caption can be misclassified. The models often failed to capture the nuanced semantic relationship between the image and text, leading to incorrect predictions. Moreover, as shown in Appendix~\ref{sec:appendix_hm_examples}, Figure~\ref{fig:hatefumemes_ea_appendix}, our analysis indicates that existing embedding approaches do not counter the exploitation of unimodal priors and hence do not address the change in the semantics of the meme.
These failure modes suggest fundamental limitations in current fusion architectures that need to be addressed through more sophisticated approaches.

\begin{table*}[!t]
\centering
\renewcommand{\arraystretch}{1.5}
\resizebox{\textwidth}{!}{%
\begin{tabular}{|c|c|c|c|c|}
\hline
  \textbf{Video Description} &
  \textbf{Modality} &
  \begin{tabular}[c]{@{}c@{}}\textbf{MO-Hate} $\rightarrow$ (BCD1) \\ \textbf{BART + CLAP + DINOv2}\end{tabular} &
  \begin{tabular}[c]{@{}c@{}}\textbf{Baseline} $\rightarrow$ (HCC1) \\ \textbf{HXP + CLAP + CLIP (Concat)}\end{tabular} \\ \hline
  \begin{tabular}[c]{@{}c@{}}Video contains anime and video subtitles\\ that do not match the audio. Audio contains \\ hate speech, subtitles and visuals are misleading.\end{tabular} &
  \begin{tabular}[c]{@{}c@{}}Text + \\ Audio \end{tabular} &
  \begin{tabular}[c]{@{}c@{}}Pred Label: 1, True Label: 1 \\ Correctly classified the video \\ utilizing audio input.\end{tabular} &
  \begin{tabular}[c]{@{}c@{}}Pred Label: 1, True Label: 1 \\ Correctly classified video \\ utilizing audio input.\end{tabular} \\ \hline
  \begin{tabular}[c]{@{}c@{}}Video contains a cartoon while the audio \\ contains hate speech repeating the word \textit{n*gg**}.\end{tabular} &
  \begin{tabular}[c]{@{}c@{}}Audio + \\ Text\end{tabular} &
  \begin{tabular}[c]{@{}c@{}}Pred Label: 0, True Label: 1 \\ Unable to correctly classify video \\ due to the partial utterance of slur.\end{tabular} &
  \begin{tabular}[c]{@{}c@{}}Pred Label: 0, True Label: 1 \\ Incorrectly classified video \\ due to partial utterance of slur.\end{tabular} \\ \hline
  \begin{tabular}[c]{@{}c@{}}Video contains hateful symbols displayed \\ throughout the video along with some sound.\end{tabular} &
  \begin{tabular}[c]{@{}c@{}} Visual \end{tabular} &
  \begin{tabular}[c]{@{}c@{}}Pred Label: 0, True Label: 1 \\ Without text data, the model \\ can only learn from the video frames.\end{tabular} &
  \begin{tabular}[c]{@{}c@{}}Pred Label: 0, True Label: 1 \\ Without text data, model \\ can only learn from the video frames.\end{tabular} \\ \hline
  \begin{tabular}[c]{@{}c@{}}Video shows violence and physical \\ altercation, but, there is no hate speech.\end{tabular} &
  \begin{tabular}[c]{@{}c@{}}Text + \\ Audio + \\ Visual \end{tabular} &
  \begin{tabular}[c]{@{}c@{}}Pred Label: 0, True Label: 0 \\ Correctly classified video\\ as non-hateful.\end{tabular} &
  \begin{tabular}[c]{@{}c@{}}Pred Label: 0, True Label: 0 \\ Correctly classified video\\ as non-hateful.\end{tabular} \\ \hline
  \begin{tabular}[c]{@{}c@{}}Video shows picture of a cop restraining a \\ person; no explicit signs of hate speech.\end{tabular} &
  Visual &
  \begin{tabular}[c]{@{}c@{}}Pred Label: 0, True Label: 0 \\ Correctly classified as non-hateful \\ even though picture looks aggressive.\end{tabular} &
  \begin{tabular}[c]{@{}c@{}}Pred Label: 1, True Label: 0 \\ Incorrectly classified as hate speech \\ due to aggressive-looking visuals.\end{tabular} \\ \hline
\end{tabular}%
}
\caption{Examples of a few hate and non-hate videos along with their description. The modality that could be used to predict them has been provided. In addition, the predictions for both \textbf{HCC1} and \textbf{BCD1} models and the likely explanation are provided.}
\label{hatemm_error_analysis_table}
\end{table*}

\section{Conclusion and Future Work}
This paper has investigated the challenging problem of multimodal hate speech detection, with a specific focus on the performance discrepancies between video-based and image-text-based content. We conducted a systematic analysis of fusion-based approaches, evaluating their effectiveness on two distinct datasets; HateMM~\citep{das2023hatemm} for video content and the Hateful Memes Challenge dataset (HMC)~\citep{kiela2020hateful} for image-text combinations. 
The effectiveness of fusion approaches varies dramatically depending on the modality combination. Simple embedding fusion achieves state-of-the-art performance on the video-based HateMM dataset, demonstrating the potential of leveraging synchronised multimodal information.
As the video offers three separate types of modalities  - audio, keyframes from the video, and text transcription, we found that a simple embedding fusion of each of the three provided an effective representation. This is because, video benefits from the temporal aspects where the richness of information provides a lot of context from which analysis of video content can benefit. The majority of the videos in the dataset were short and choosing only $100$ frames did not affect the result since, upon manual inspection, the longer videos were deemed to be non-hateful.

However, on the HMC dataset, our experiments demonstrated that both simple fusion and modality order-aware fusion approaches fail to fully capture the nuanced semantic relationships in image-text pairs. If the hate is implicit in the text then determining when a meme is hateful will only be detectable if combined with the image. The same argument holds when the hate is implicit in the image and the overall meme will only be hateful when combined with the text. This is made more complex when the sets of images and text include ``benign confounders''. We have shown that individual embeddings of either the image or text and then by the application of simple fusion we cannot achieve robust results. This is because of the prior arguments of text and image information being interrelated. Moreover, the interplay between the text and the image is not being utilised in the individual representation prior to fusion, and this is the main limitation of the current approaches.

In the future, we propose the development of a unified framework that incorporates improved fusion and embedding extraction techniques to handle the complexities of meme-based hate speech. This framework could include components for object detection, captioning, and visual question answering to better understand the context and meaning of different modalities. 
For the HateMM dataset, we aim to enhance the framework's ability to identify the timestamp range for hateful content in videos. This will involve training models on datasets that provide timestamp annotations indicating hateful segments. Additionally, we will work on improving the dataset itself by curating more instances and ensuring that videos have manually transcribed, high-quality text captions to mitigate the issues caused by inaccurate ASR. Further research is needed to address the practical challenges of deploying multimodal hate speech detection systems in real-world settings. This includes addressing issues such as computational efficiency, scalability, and the need for continuous monitoring and updates.
By pursuing these research directions, we believe that it is possible to develop more robust, accurate, and practical multimodal hate speech detection systems that can effectively combat the spread of hate speech online.

\section{Limitations and Social Impact}

While our work presents a comprehensive evaluation of existing approaches on varying modalities for multimodal hate speech detection, certain limitations should be acknowledged. The datasets used in our experiments, HateMM and HMC may contain inherent biases that could influence the model's performance. For instance, the data collection process or the choice of data sources (\textit{e.g.,} specific social media platforms) could introduce biases related to demographics, or topic distributions. Such biases can affect the generalisability of our findings to other datasets or real-world scenarios which potentially has a societal impact. While the models that we trained using the Baseline and MO-Hate architectures reach SoTA scores on the HateMM dataset, they perform significantly worse on the HMC dataset. This shows the need for sophisticated information extraction from the modalities and a robust fusion mechanism. Our evaluation primarily focuses on standard classification metrics, such as F1 and AUROC scores. However, these metrics may not fully capture the complexities and nuances involved in multimodal hate speech detection in a real-world scenario.

\bibliographystyle{ACM-Reference-Format}
\balance
\bibliography{acm_wc_2025}

\appendix

\section{Transcription Quality: Vosk (vosk-model-en-us-0.22) vs. Whisper (tiny)}

\label{sec:appendix_tq}

\begin{figure}[!ht]
    \centering
    \includegraphics[width=\columnwidth]{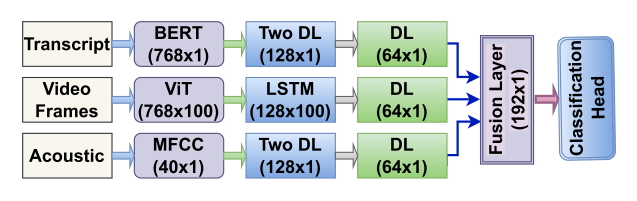}
    \caption{Simple Fusion architecture as shown in \citet{das2023hatemm}.}
    \label{fig:baseline}
\end{figure}

\section{Other Experimental Results}
\label{sec:appendix_results}

\begin{figure*}[!t]
    \centering
    \includegraphics{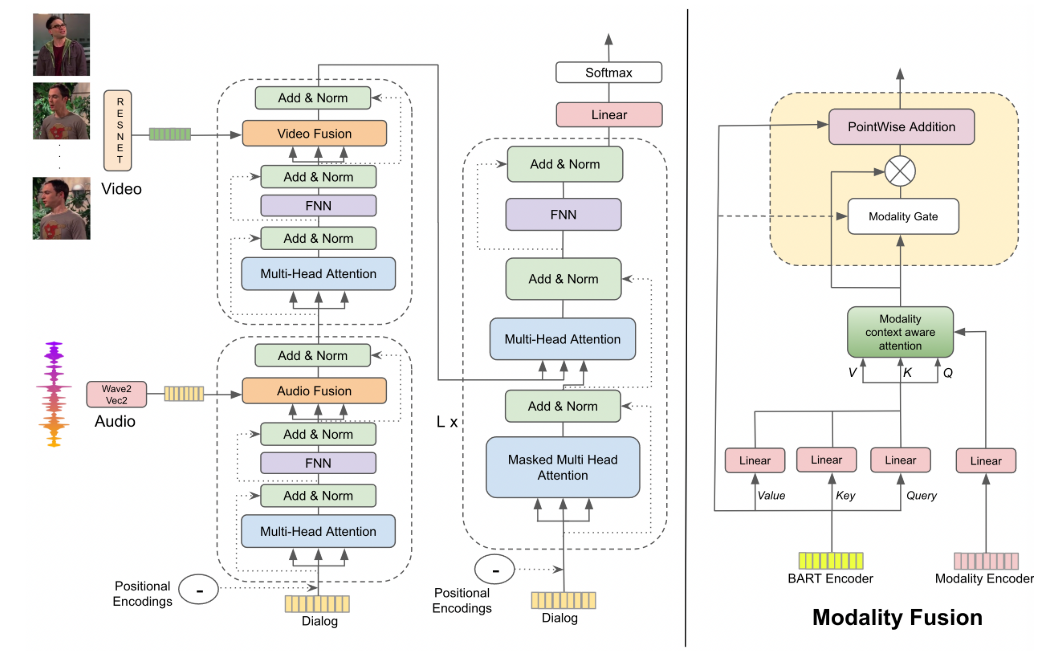}
    \caption{MO-Hate architecture as shown in \citet{tomar2023your}.}
    \label{fig:mo-hate}
\end{figure*}

\begin{table*}[!ht]
    \centering
    \renewcommand{\arraystretch}{1.5}
    \begin{adjustbox}{max width=\textwidth}
    \begin{tabular}{c|c c c c c}
        \toprule
        & \textbf{Models} & \textbf{F1 (M)} & \textbf{P (M)} & \textbf{R (M)} & \textbf{Acc} \\
        \midrule
        \multirow{3}{*}{\rotatebox{90}{Simple Fusion}} 
        & HXP + Wav2Vec2 + DINOv2 LHS (Concat - EF) & 0.760 & 0.755 & 0.764 & 0.801 \\
        & HXP + WAV2VEC2 + ViT (EW Product - EF) & 0.757 & 0.762 & 0.753 & 0.808 \\
        & HXP + MFCC + ViT (Weighting Technique - LF) & 0.819 & 0.840 & 0.800 & 0.854 \\
        & HXP + MFCC + ViT (Trained Probabilities - LF) & 0.747 & 0.765 & 0.729 & 0.803 \\
        & HXP + AVGG19 + ViT (Concat - EF) & 0.680 & 0.784 & 0.600 & 0.773 \\
        & BERT + AVGG19 + ViT (Concat - EF) & 0.569 & 0.695 & 0.482 & 0.707 \\
        \midrule
        \multirow{3}{*}{\rotatebox{90}{MO-Hate}} 
        & BART + ViT + Wav2Vec2 & 0.820 & 0.822 & 0.822 & 0.822 \\
        & BART + ViT + CLAP & 0.818 & 0.821 & 0.817 & 0.817 \\
        & BART on whisper + CLAP + CLIP pooled & 0.789 & 0.790 & 0.791 & 0.791 \\
        & Visual (ViT) + Audio (Wav2Vec2) + Text (BART) & 0.752 & 0.763 & 0.761 & 0.761 \\
        \bottomrule
    \end{tabular}
    \end{adjustbox}
    \caption{EW: element-wise, EF: early fusion, LF: late fusion methods as described in \citet{sai2022explorative}. Models starting with BART in MO-Hate are text-first models whereas, Visual (ViT) is the visual-audio-text fusion model in that order. M: macro, P: precision, R: recall, Acc: accuracy.}
    \label{hatemm_table_appendix}
\end{table*}

\citet{das2023hatemm} used the Vosk tool in their experiments to transcribe the audio and use the text for embeddings. However, when we experimented with Whisper, there was a significant improvement in the quality of the transcript and thus we decided to use it for all the audio data. Here are a few samples on which Whisper performs well:

\vspace{0.5cm}
\textbf{Video ID}: 'hate-video-32.mp4'

\textbf{Vosk transcript}: "the the the the the the the the the the"

\textbf{Whisper transcript}: "I'm going to go home. I'm not sure what you're doing, so I'll know what I can. One day I'll get, and much worse than to be. I thought I was about to land, so stuff that's so thick and I'm not going to lay down and all the sleep. Well, I'd be glad to lay down with all the sleep. When I wake up and I've been 12 and 13, I'll be able to stay out of the house and I'll stop again. And you've got down there, Gail. You know what I mean? I want to wait down there and try and get along with that and couldn't get no ladder. I want you to do that. I'm not that black and I'll stop it up the way and I'll sit now, you're still in. Well, I did say that, nigga ain't never spoke to me tonight, there you are. I know it's male, black man nigga done to this. I'm feeling like a big man, nigga done to this. Still out there, I'm still out there, I'm still out there, I'm still out there, I'm still out there, I'm not gonna have to be."

\vspace{0.5cm}
\textbf{Video ID}: 'hate-video-308.mp4'

\textbf{Vosk transcript}: "when mexico sends its people they're not sending their best they're not sending you the the the it's worth the read the the the mexicans read the the the the the"

\textbf{Whisper transcript}: "When Mexico sends his people, they're not sending their best. They're not sending you. They're sending people that have lots of problems. And they're bringing those problems with them. They're bringing drugs, they're bringing crime, they're rapping at some. Are you sure my little people? It seems that deportations needed A big ball to be made Good good advice, a Trump unheated The borders are sure red The right needs a new direction Now the old global is the way It's a long global to correction To protect our DNA And so can server this worse burn And all the blue blue wings brought down And all the Mexicans returned On that day, keep streaming, bash the nation Cheers are burning, I'm so strong Keep streaming, bash fashionation, white's a learning, marching out. Well the truth it got to be a region, multiculturalism so. And we need to see that there's no lapse in in We've driven the blindfold And so conservatives were spurned And all their influence brought down And all the Mexicans returned On that day, he extreme in bash the nation, jes of burning, I'm so strong, keep Until our country's representatives can figure out what the hell is going on. And so conservatives were spurned. And all their influence brought down. And all the Mexicans re-zurned. I'm so strong Keep streaming, fashion nation What's our learning? Marching out Keep streaming, fashion nation"

\vspace{0.5cm}
\textbf{Video ID}: 'hate-video-430.mp4'

\textbf{Vosk transcript}: "the why can't the the the hey the the the yeah the asked the the the the the the the"

\textbf{Whisper transcript}: "And another on the right to car No I can't get those from us And I might have been good since guns What main camp I race will we found weapons And I've got to lose moving on my own And now so many my tags I will not enable for his dust And let's be clear not just no one You did not break me I'm worth fighting for peace We love God, why it's skin and the passion's heart But your edgeiness was made to shine And take the dunes and do better hard Yeah, I miss that they've'm a stepping block together Free things should be a I'm a news I can't abide And I know that that I can survive Cause shots were falling, saved my life The media fucked up my life So bad I'm doing everything I can But when I ground with the dust I got the ire of the chosen bonds You did not break me I'm still carrying my keys Well I've got what's getting the best you've had But your ageiness is way too short We'll take the pill, don't lose until you make the heart Behind me, step and pop together We say it should be a on Cause I've got white skin that affect your heart But you're ready to just make some sure I'll take the deal, don't listen till you make the heart Yeah, I just took it, pump again, I'll raise it, shoot it, yeah Who was I ever? She's hard, all the reasons she's here, all I was, because I ever, she's hard It's time"

\section{Simple Embedding Fusion and MO-Hate Architectures}

\label{sec:appendix_arch}

Figure~\ref{fig:baseline} shows the Simple Embedding Fusion architecture and Figure~\ref{fig:mo-hate} shows the MO-Hate architecture.

Table~\ref{hatemm_table_appendix} presents additional results on the HateMM dataset where EW indicates an element-wise product was computed for embedding fusion. Further, EF indicates early fusion whereas LF indicates a a late fusion approach was utilised. Similarly, Table~\ref{hatememes_table_appendix} presents additional results on the HMC dataset using similar approaches.

\begin{table*}[!ht]
    \centering
    \renewcommand{\arraystretch}{1.5}
    \begin{adjustbox}{max width=0.9\textwidth}
    \begin{tabular}{c|c c c c c c}
        \toprule
        & \textbf{Models} & \textbf{AUROC} & \textbf{F1 (M)} & \textbf{P (M)} & \textbf{R (M)} & \textbf{Acc} \\
        \midrule
        \multirow{3}{*}{\rotatebox{90}{Simple Fusion}} 
        & BERT + ViT (Trained Probabilities - LF) & 0.557 & 0.407 & 0.516 & 0.336 & 0.595 \\
        & BERT + ViT (Weighting Technique - LF) & 0.527 & 0.221 & 0.537 & 0.139 & 0.594 \\
        & HXP + ViT (Concat - EF) & 0.510 & 0.225 & 0.449 & 0.150 & 0.572 \\
        \midrule
        \multirow{2}{*}{\rotatebox{90}{\small MO-Hate}} 
        & BART + ViT & 0.606 & 0.590 & 0.632 & 0.611 & 0.611 \\
        & Image (ViT) + Text (BART) & 0.504 & 0.448 & 0.508 & 0.511 & 0.511 \\
        \bottomrule
    \end{tabular}
    \end{adjustbox}
    \caption{Model performance on the HMC dataset using different ensembles. EF: early fusion, LF: late fusion methods as described in \citet{sai2022explorative}. All MO-Hate models are fused in the order mentioned. M: macro, P: precision, R: recall, Acc: accuracy.}
    \label{hatememes_table_appendix}
\end{table*}

\section{A few more Hateful Memes examples}

\label{sec:appendix_hm_examples}

\begin{figure*}[!ht]
    \centering
    \begin{subfigure}{0.4\textwidth}
        \centering
        \includegraphics[width=7cm]{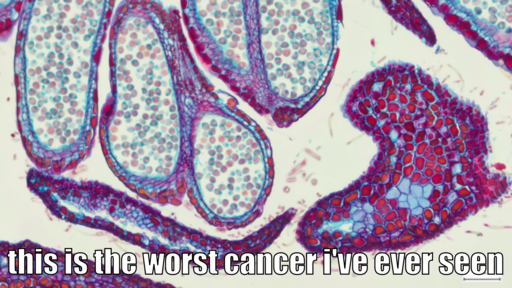}
        \caption{True Label: 0 Pred Label: 0}
        \label{fig:meme3}
    \end{subfigure}
    \hfill
    \begin{subfigure}{0.4\textwidth}
        \centering
        \includegraphics[width=7cm]{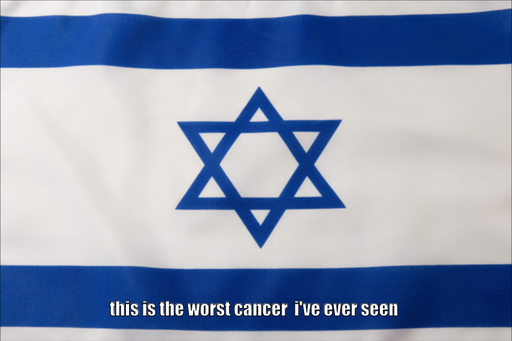}
        \caption{True Label: 1 Pred Label: 0}
        \label{fig:meme4}
    \end{subfigure}
    \vfill
    \begin{subfigure}{0.4\textwidth}
        \centering
        \includegraphics[width=7cm]{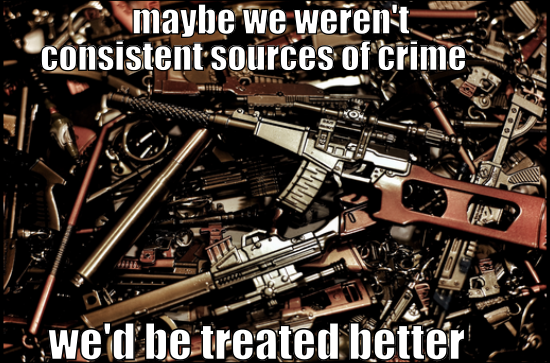}
        \caption{True Label: 0 Pred Label: 0}
        \label{fig:meme5}
    \end{subfigure}
    \hfill
    \begin{subfigure}{0.4\textwidth}
        \centering
        \includegraphics[width=7cm]{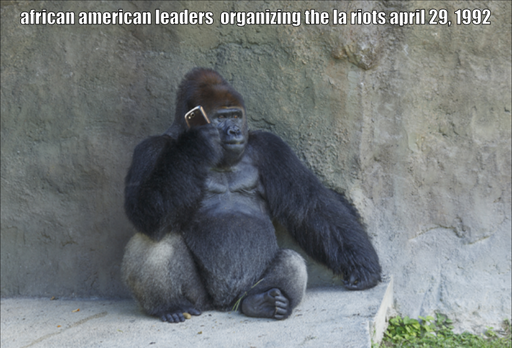}
        \caption{True Label: 1 Pred Label: 0}
        \label{fig:meme6}
    \end{subfigure}
    \caption{More predictions and true labels for some of the HMC instances using the \textbf{MBD1} model.}
    \label{fig:hatefumemes_ea_appendix}
\end{figure*}

Figures~\ref{fig:meme3} and~\ref{fig:meme4} are image confounders and the prediction results show that the \textbf{MBD1} fails to recognise the hateful images irrespective of the type of confounder. This is also visible in figures~\ref{fig:meme5} and~\ref{fig:meme6}.

\end{document}